\def\year{2016}\relax
\documentclass[letterpaper]{article}
\usepackage{aaai16}
\usepackage{times}
\usepackage{helvet}
\usepackage{courier}
\usepackage{url}
\usepackage{latexsym}
\usepackage[utf8]{inputenc} 
\usepackage{graphicx}
\usepackage{amsfonts}
\usepackage{amsmath}
\usepackage{amssymb} 
\usepackage[usenames,dvipsnames]{color}
\usepackage{enumitem}
\usepackage{array}
\usepackage{subcaption}
\usepackage{multirow}

\newcommand{\newcite}[1]{\citeauthor{#1} \shortcite{#1}}
\begin{document}
%
\title{Expanding Subjective Lexicons for Social Media Mining \\with Embedding Subspaces}


\author{Silvio Amir, Rámon Astudillo, Wang Ling$^{*}$, Paula C. Carvalho, Mário J. Silva\\
INESC-ID Lisboa, Instituto Superior Técnico, Universidade de Lisboa\\
$^{*}$ Google DeepMind, London \\ 
{samir@inesc-id.pt~ramon@astudillo.com~lingwang@google.com~pcc@inesc-id.pt~mjs@inesc.id.pt}\\
}

\maketitle
\begin{abstract}

Recent approaches for sentiment lexicon induction have capitalized on pre-trained word embeddings that capture latent semantic properties. However, embeddings obtained by optimizing performance of a given task (e.g. predicting contextual words) are sub-optimal for other applications. In this paper, we address this problem by exploiting \textit{task-specific} representations, induced via embedding sub-space projection. This allows us to expand lexicons describing multiple semantic properties. For each property, our model jointly learns suitable representations and the concomitant predictor. Experiments conducted over multiple subjective lexicons, show that our model outperforms previous work and other baselines; even in low training data regimes. Furthermore, lexicon-based sentiment classifiers built on top of our lexicons outperform similar resources and yield performances comparable to that of supervised models.


\end{abstract}

\section{Introduction}

The rise of the social web brought about an unprecedented volume of data about human interactions, paving the way for a new, computational, approach to social sciences. Nowadays, scholars are able to study societal and behavioral dynamics through the analysis of large-scale social networks and vast amounts of social media. For example, natural language processing techniques have been applied to massive microblogging repositories to investigate a wide range of phenomena, such as population well-being~\cite{Mitchell2013}, political participation~\cite{tumasjan2010predicting} and public opinion~\cite{O'Connor2010}. 

One of the key resources to support this kind of analyses are subjective lexicons---i.e., lists of words with semantic annotations. Particularly, \textbf{sentiment} lexicons, which categorize words according to the polarity of sentiment they convey; and \textbf{emotion} lexicons, which quantify the emotional states or responses evoked by a given word (e.g. \textit{joy} or \textit{arousal}). Typically, these resources are manually created by experts or via crowdsourcing campaigns; a process that can become expensive and time-consuming. Therefore, manually crafted lexicons are necessarily incomplete, thus failing to capture the use of non-conventional word spellings, slang and new expressions commonly found in social media. 

The automatic extraction of lexicons is a well-known and widely studied problem. Most proposed solutions are predicated on the idea that \textit{similar} words should have similar labels. These solutions differ, essentially, along two axes. First, how the word similarities are captured, e.g. leveraging knowledge bases, such as WordNet~\cite{hu2004mining} or from word co-occurrence statistics derived from corpora analysis~\cite{Bestgen12Checking}; and second, how the label assignment is operationalized, e.g. with supervised classifiers~\cite{esuli2006sentiwordnet} or using graph-based label propagation algorithms~\cite{rao2009semi}. 

Recent work has also begun exploring neural word embebddings, due to their ability to capture word similarities and latent semantic properties. \newcite{tang-EtAl2014} induced Twitter sentiment lexicons using \textit{sentiment-specific} word embeddings, obtained via distant supervision, whereas \newcite{Amir} proposed a predictive model, leveraging unsupervised embeddings as features. Using this approach, \citeauthor{Amir} developed the top ranking submission of a lexicon expansion shared task organized by SemEval 2015~\cite{SemEval15Task10}. However, both these methods are inherently limited by their choice of word representations. On the one hand, \citeauthor{tang-EtAl2014}'s approach uses embeddings tailored to capture sentiment information, but then it can only be used for polarity lexicons. \citeauthor{Amir}'s method, on the other hand, can be used for other lexicon types, but it uses generic, unsupervised embeddings which are sub-optimal for specific downstream models~\cite{astudillo-EtAl:2015:ACL-IJCNLP,labutov2013re}. 

In this paper, we present an approach to overcome these limitations. Following \citeauthor{Amir}, we expand lexicons for social media mining with predictive models, leveraging unsupervised word embeddings features. This allows us to deal with lexicons describing different properties. Unlike their approach, however, we induce and exploit intermediate, \textit{task-specific} representations via embedding subspace projection~\cite{astudillo-EtAl:2015:ACL-IJCNLP}. The evaluation was conducted over seven lexicons describing 15 subjective properties. The results show that our models largely outperform the other baselines (with two exceptions). To assess the quality of our lexicons, we built and evaluated lexicon-based Twitter sentiment classifiers. We found that our lexicons: (i) outperform other similar resources; (ii) yield performances comparable to that of supervised models.

\section{Related Work}
\label{sec:related}

The previous work on automatic lexicon induction, can be roughly divided into two classes: \textit{knowledge-based} and \textit{corpora-based}. These approaches assume that there is a small number of words, for which the labels are known (sometimes referred as the \textit{seed set}). Knowledge-based approaches, then use lexical databases such as WordNet to, e.g., exploit word relations such as antonymy/synonymy or hyponymy/hypernymy between new words and words in the seed set~\cite{hu2004mining,kim2006identifying,rao2009semi}. Others, classify new words leveraging distances to known words in the synset graph~\cite{kamps2004using}. However, these methods are unsuited for the social web since they rely on formal lexical resources that do not encompass the informal language and writing style typical of this domain. 

Corpora-based approaches, implicitly exploit the distributional hypothesis, which works under the assumption that similar words tend to occur in similar \textit{contexts}~\cite{harris1954distributional}. Based on this idea, word similarities can be computed from a term co-occurrence matrix, built from large corpora, using e.g. the \textit{point-wise mutual information} (PMI) between terms~\cite{Turney03Measuring,kiritchenko2014sentiment} or with vector distance metrics, over the space induced by \textit{Latent Semantic Analysis} \cite{Bestgen12Checking,Yu:2013:UCE:2438098.2438152}. 

In the last few years, efficient neural language models have been proposed to induce word embeddings (i.e. dense feature vectors) by learning to predict the surrounding contexts of words in large corpora(e.g. ~\cite{pennington2014glove}). Recent work on lexicon expansion has also explored these representations, due to their ability to capture word similarities and latent semantic properties~\cite{tang-EtAl2014,Amir,rothe2016ultradense}. Nevertheless, because these vectors are estimated by minimizing the prediction errors made on generic, unsupervised tasks, they can be sub-optimal for specific downstream models. In this work, we take this observation into account and demonstrate that better models can be induced, by jointly learning \textit{task-specific} representations and predictors.

\section{Learning Task-Specific Embeddings}

Neural embeddings are \textit{distributed representations}, i.e. each concept is described by multiple \textit{features} (vector dimensions can be interpreted as abstract features) and each feature can be involved in describing multiple concepts \cite{hinton1986learning}\footnote{On the other hand, \textit{symbolic} representations associate each feature to only one concept---e.g. the \textit{one-hot} encoding, represents word $i$ as a zero vector with the value 1 on the $i$-th dimension.}. However, we hypothesize that not all features contribute equally to capture any given aspect of the input. Thus, if we knew exactly which subset of features describe a given property (e.g. sentiment), we could extract a more compact representation containing only the meaningful information. This would eliminate noise and irrelevant aspects of the data. But more importantly, predictors based on smaller representations require fewer free parameters, which makes them easier to train with small datasets without overfiting.


One common solution to extract compact representations from a large feature space, is to perform dimensionality reduction by Principal Component Analysis (PCA). But this is a generic linear transformation that does not take into account the prediction targets, hence it is sub-optimal\footnote{Moreover, word embeddings can be seen as the result of a dimensionality reduction over a sparse word-context co-occurrence matrix (see \cite{Baroni14Dont} and \cite{pennington2014glove}). Thus it is not clear how to interpret the result of a subsequent dimensionality reduction.}. Alternatively, we could use the full embeddings and fit a predictor with a sparsity inducing objective, e.g. using $\ell_1$-norm regularization~\cite{tibshirani1996regression}. This regularizer tries to bring the weights associated to some input dimensions down to zero, essentially eliminating the contribution of some features. However, without any prior knowledge about the structure of the embeddings there is no guarantee that only the irrelevant dimensions would be eliminated. Moreover, given the distributed nature of word embeddings, simply ignoring arbitrary dimensions may degrade their expressiveness. Of course, we could just use vectors with fewer dimensions but small embeddings have less capacity and tend to be better suited to syntactic tasks than to semantic ones~\cite{ling-EtAl:2015:EMNLP2}. 

\subsection{Word Embedding Sub-spaces}

A simple, yet effective solution to this problem consists of estimating linear projections from generic embeddings to lower-dimensional \textit{sub-spaces}~\cite{astudillo-EtAl:2015:ACL-IJCNLP}. Given an embedding matrix $\mathbf{E} \in \mathbb{R}^{d \times |\mathcal{V}|}$, where the columns represent words from a vocabulary $\mathcal{V}$ and $d$ is the embedding size, one can induce new representations with a factorization of the input as $\mathbf{S} \cdot \mathbf{E}$ where $\mathbf{S} \in \mathbb{R}^{s \times d}$,  with $s \ll d$, is a (learned) linear projection matrix.  This is similar to PCA, but here the projection is estimated to directly optimize the prediction of the target labels. Therefore, the transformation corresponds to adapting generic, unsupervised representations into task-specific ones. The intuition is that by aggressively reducing the representation space, the model is forced to learn only the most discriminative aspects of the input with respect to the prediction targets\footnote{This is also related to the idea of learning representations with information bottlenecks, e.g with auto-encoders~\cite{hinton2006reducing}}.

\section{Proposed Approach}

As we discussed above, lexicons can be expanded under the assumption that similar words should have similar labels. To operationalize this assumption, we capitalize on two fundamental properties of neural word embeddings: first, they encode functional (i.e., semantic and syntactic) similarities in terms of geometric locality. This will allow us to predict consistent labels to inflections, spelling variations and synonyms of a word; and second, they capture latent word aspects, some of which correspond to subjective properties, e.g. sentiment polarity.

Our approach to lexicon expansion consists of training models to predict the labels of pre-existing lexicons, leveraging unsupervised word embeddings as features. We further assume that different aspects captured by these embeddings are encoded in some (unknown) subset of features. Therefore, we adopted \newcite{astudillo-EtAl:2015:ACL-IJCNLP} Non-Linear Subspace Embedding model (\textsl{NLSE}), to jointly learn embedding sub-space projections that better represent specific word properties, and the concomitant predictors. 

The NLSE is essentially a feed-forward neural network, with a single hidden layer and a factorization of the input layer. Denoting a lexicon of $n$ word/label pairs as ${\mathcal{D} = \{(w_1,y_1),\dots ,(w_n,y_n)\}}$ where $y$ is a categorical random variable over a set of classes $\mathcal{Y}$, the model estimates the probability of each possible category $y=k \in \mathcal{Y}$ given a word $w_i$ as
\begin{align}
p(y=k| w_i;\theta) &= \mathrm{softmax}(\mathbf{h}_i) =\frac{e^{\mathbf{W}_k \cdot \mathbf{h}_i}}{\sum_{j=1}^{\mathcal{Y}}{e^{\mathbf{W}_j \cdot \mathbf{h}_i}}}\label{eq:nlse}\\
\nonumber \mathbf{h}_i &= \sigma\left(\mathbf{S}\cdot \mathbf{E}_{[i]} \right)
\end{align}
Here, $\mathbf{E}_{[i]}$ selects the $i$-th column of the embedding matrix (corresponding to word $w_i$), $\mathbf{h}_i \in [0, 1]^{s}$ is a vector of activations computed by the hidden layer and $\sigma(\cdot)$ denotes an element-wise sigmoid non-linearity. The matrix $\mathbf{W} \in \mathbb{R}^{|\mathcal{Y}| \times s}$ maps the embedding sub-space to the classification space. The parameters $\theta=\{\mathbf{W,S}\}$ are estimated to minimize the inverse log-likelihood of the training data
\begin{align}
\theta \leftarrow \underset{\theta}{\min} - \sum_{(w_i,y_i) \in \mathcal{D}} \log p(y_i|w_i;\theta)
\label{eq:nlse_loss}
\end{align}
The original embedding matrix is kept fixed, while the projection parameters are estimated jointly with the predictor, thus the model induce and exploit a compact, task-specific, representation that preserves the rich information captured by the embeddings.

Note that the model in Eq. \ref{eq:nlse} produces a probability distribution over the output classes, hence it is only suitable for categorical lexicons. Nonetheless, it can be easily adapted to continuous outputs by replacing the \textit{softmax} classifier with a simple linear regressor:
\begin{equation}
\hat{y}_{i} = g(w_i;\theta^{\prime}) = \mathbf{w} \cdot \mathbf{h}_i + b
\label{eq:nlse_reg}
\end{equation}
\noindent where $\mathbf{w} \in \mathbb{R}^s$ and $b \in \mathbb{R}$ are the regression weights and bias, respectively. The parameters $\theta^{\prime}=\{\mathbf{w},b,\mathbf{S}\}$ are estimated by minimizing the Mean Squared Error over the training data
\begin{equation}
\theta^{\prime} \leftarrow \underset{\theta^{\prime}}{\min} \sum_{(w_i,y_i) \in \mathcal{D}} (y_i-\hat{y}_i)^2
\label{eq:nlse_reg_loss}
\end{equation}
\noindent The loss functions in Eq. \ref{eq:nlse_loss} and Eq. \ref{eq:nlse_reg_loss} can be minimized with standard gradient based optimization methods. After training, the models can be employed to extrapolate the labels of any word for which an embedding is available. Furthermore, by using embeddings induced from social media, we can adapt any pre-existing lexicon to include terms that are used in this domain. 


\section{Predicting Lexicon Labels}

In this section, we evaluate our method at inferring different subjective word properties, namely the \textbf{sentiment polarity}, \textbf{happiness level}, \textbf{affective responses} (specifically, the valence, arousal and dominance) and \textbf{emotion association} (concretely, \newcite{Plutchik}'s basic emotion set). We trained the models described in the previous section to predict the labels assigned by human judges to several well-known subjective lexicons. These are summarized in Table \ref{table:lexicons}, and consist of two groups: categorical lexicons, associating words to specific classes e.g., \textit{positive polarity} (on the top rows); and real-valued lexicons, assigning continuous values to words e.g., \textit{valence} (bottom rows). 

\begin{table}[!h]
\begin{center}
\scalebox{0.8}{
\begin{tabular}{l r l}
 & {\bf \# Words }\\ 
\hline\hline
\textbf{Opinion Mining Lexicon (OML)}~\cite{hu2004mining}  & 6,787 \\ 
\textbf{MPQA}~\cite{wilson2005recognizing}  & 6,886 \\ 
\textbf{Emotion Lexicon (EmoLex)}~\cite{mohammad2013a}  & 14,174 \\ 
\hline
\textbf{ANEW}~\cite{bradley1999affective}        & 1,040  \\
\textbf{SemEval (Sem-Lex)}~\cite{SemEval15Task10} & 1,515 \\
\textbf{LabMT}~\cite{dodds2011temporal}          & 10,000 \\
\textbf{Ext-ANEW}~\cite{warriner2013norms}  & 13,915 \\ 
\end{tabular}
}
\end{center}
\caption{Lexicons}
\label{table:lexicons}
\end{table}
\subsection{Experimental Setup}

Our approach requires an unlabeled corpus to support the induction of the word embedding matrix. Following \newcite{Amir}, we induced 600 dimensional \textsl{Structured Skip-gram} word embeddings~\cite{ling-EtAl:2015:EMNLP2}. We evaluated several baselines utilizing the same word embeddings as the input, but with predictors based on three variants of Support Vectors Machines (\textsl{SVM}) and Support Vectors Regression (\textsl{SVR})~\cite{vapnik2000nature}: (i) \textbf{linear}; (ii) with \textbf{$\ell_1$-norm regularization}; and (iii) with \textbf{non-linear kernel}. For the latter, we used a radial basis function (RBF) kernel of the form ${\mathrm{k}(\mathbf{x}_i,\mathbf{x}_j) = e^{(-\gamma |\mathbf{x}_i-\mathbf{x}_j|^2)}}$ with $\gamma > 0$, where $\mathbf{x}$ denotes a feature vector. In this case, the models learn a linear function in the space induced by the kernel and the data, which corresponds to a non-linear function in the original space. This baseline corresponds to \newcite{Amir} model. Finally, we considered the linear models but using compact representations obtained with PCA.

The experiments were performed by splitting the labeled data (i.e., the lexicons) in 80\% for model training and the remaining 20\% for evaluation. Then, for each experiment, 20\% of the training data was reserved for hyper-parameter tuning via grid-search. We tuned the misclassification cost in the range ${C=[1e^{-2}, 1^{e-1}, 1, 10,50,100,150]}$ for all the SVM and SVR models. Furthermore, we searched over the following, model specific, hyper-parameters: kernel widths, in the range ${\gamma=[1e^{-3}, 1e^{-2}, 1^{e-1}, 1, 10]}$ for the RBF kernel baselines; regularization constant in the range ${\lambda=[1e^{-4}, 1e^{-3}, 1e^{-2}, 1^{e-1}, 1, 10]}$ for the regularized baselines; and the number of components to keep in the PCA baselines, in the range ${S=[3, 5,10,15,20]}$. Regarding the NLSE model, we optimized the subspace size and the learning rate over the ranges ${S=[3, 5,10,15,20]}$ and ${\alpha=[1e^{-3}, 1^{e-2}, 5^{e-2},1^{e-1},5^{e-1}]}$, respectively. 

\begin{figure*}[htb]
\centering
\includegraphics[width=1\linewidth]{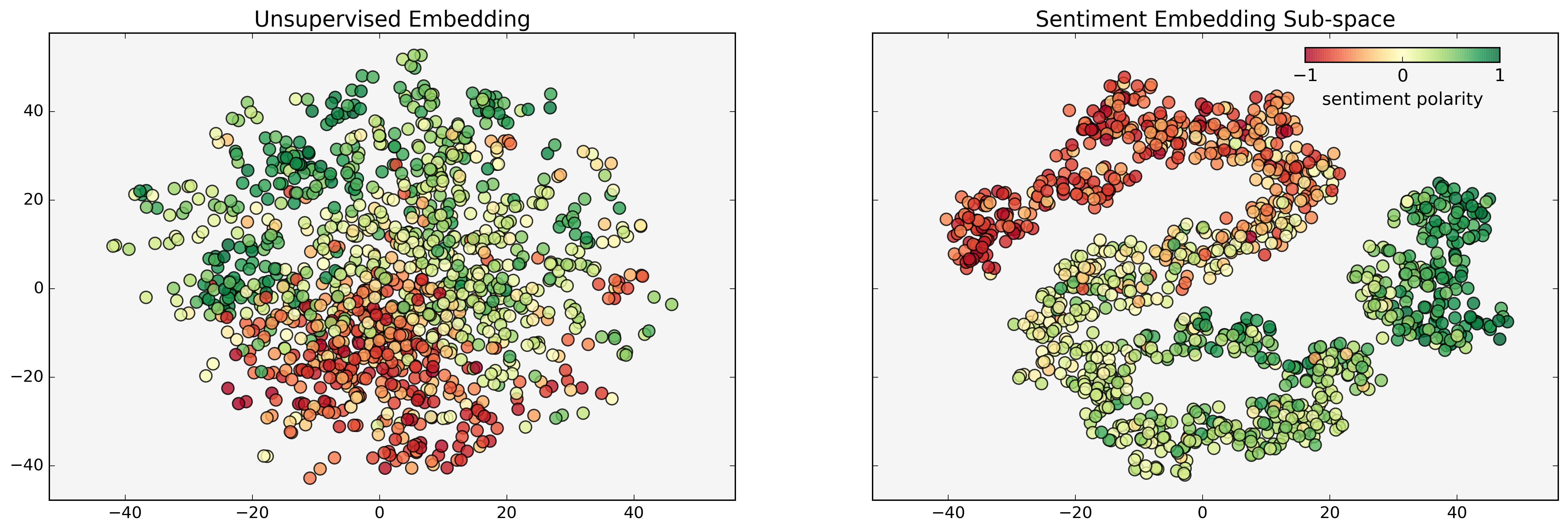}
\caption{T-SNE projection of the embeddings associated to words from \textsl{Sem-Lex}, to two dimensions. The points are colored according to their sentiment polarity. The left plot, shows the words represented as 600-dimensional unsupervised embeddings. The right plot, shows the same words represented with \textit{task-specific} embeddings induced with NLSE model.}
\label{fig:sentiment_emb}
\end{figure*}


\subsection{Results}

The main experimental results are presented in Table \ref{tab:cat}, for the categorical lexicons, and Table \ref{tab:cont}, for the continuous ones. We can see that the NLSE largely outperforms all the other baselines, apart from two exceptions, where \newcite{Amir}'s approach (RBF column) does slightly better. The results also show that the support vector models tend to perform better when used with non-linear kernels.

\begin{table}[!htb]
  \centering  
  \small
  \scalebox{0.9}{      
    \begin{tabular}{l l r | r r r r}
        & & {\bf NLSE} & \multicolumn{4}{c}{\bf SVM} \\
        & &  & {\bf linear} & {\bf$\ell_1$} & {\bf RBF} & {\bf PCA} \\
        \hline\hline            
        {\bf OML}     & sentiment    & \textbf{0.882} & 0.868 & 0.686 & 0.872 & 0.852 \\            
        \hline
        {\bf MPQA}   & sentiment    & \textbf{0.691} & \textbf{0.691} & 0.221 & 0.669 & 0.555 \\                           
                     & subjectivity & 0.825 & 0.819 & 0.798 & \textbf{0.833} & 0.805 \\   
        \hline
        {\bf EmoLex} & sentiment 	& \textbf{0.676} & 0.630 & 0.404 & 0.640 & 0.468 \\
        			 & sadness      & \textbf{0.509} & 0.340 & 0.167 & 0.334 & 0.000 \\            
					 & fear         & \textbf{0.503} & 0.373 & 0.261 & 0.394 & 0.000 \\
					 & anger        & \textbf{0.468} & 0.353 & 0.214 & 0.366 & 0.000 \\
					 & disgust      & \textbf{0.446} & 0.343 & 0.180 & 0.352 & 0.000 \\
        			 & joy          & \textbf{0.440} & 0.333 & 0.148 & 0.329 & 0.000 \\
                     & trust        & \textbf{0.403} & 0.201 & 0.190 & 0.167 & 0.000 \\                     
                     & surprise     & \textbf{0.204} & 0.167 & 0.093 & 0.119 & 0.000 \\         
                     & anticipation & \textbf{0.240} & 0.108 & 0.151 & 0.044 & 0.000 \\     
    \end{tabular}}    
    \caption{Results for categorical lexicons in terms of Avg. $F_1$ \label{tab:cat}}
\end{table} 

Regarding the baselines that try to uncover the relevant information from the embeddings (i.e., PCA and $\ell_1$), we can see that they perform very poorly. This was expected, since the former induces a low-rank approximation of the original embedding, that (tries to) preserve most of the variance. However, since word embeddings are distributed representations, the values of individual dimensions are meaningless and should be regarded as coordinates in a high-dimensional space. The latter, on the other hand, tries to drop some of the input dimensions, but in doing so degrades the information contained in the word representations. These approaches are particularly inefficient in the more nuanced properties such as fine-grained emotions. Conversely, these are precisely the cases where our approach stands-out, which underlines the benefits of inducing task-specific representations. 

\begin{table}[!htb]
  \centering   
  \small
  \scalebox{0.9}{
    \begin{tabular}{l l r | r r r r}
        & & {\bf NLSE} & \multicolumn{4}{c}{\bf SVR} \\
        & &  & {\bf linear} & {\bf$\ell_1$} & {\bf RBF} & {\bf PCA} \\        
        \hline\hline
        {\bf SemLex}   & sentiment &  \textbf{0.667} & 0.610 & 0.619 & 0.630 & 0.622 \\
        {\bf LabMT}    & happiness & \textbf{0.640} & 0.576 & 0.573 & 0.622 & 0.464 \\
        \hline        
        {\bf ANEW}     & arousal & \textbf{0.440} & 0.365 & 0.375 & 0.415 & 0.389 \\
                       & valence & \textbf{0.683} & 0.612 & 0.604 & 0.646 & 0.592 \\
                       & dominance & \textbf{0.546} & 0.477 & 0.456 & 0.494 & 0.475 \\
        \hline
        {\bf Ext-ANEW} & arousal & 0.393 & 0.373 & 0.371 & \textbf{0.397} & 0.315 \\
                       & valence &  \textbf{0.607} & 0.567 & 0.565 & 0.593 & 0.494 \\
                       & dominance & \textbf{0.480} & 0.445 & 0.443 & 0.464 & 0.405 \\
    \end{tabular}
    }          
    \caption{Results for continuous lexicons in terms of Kendall $\tau$ rank correlation \label{tab:cont} }
\end{table}%

To further illustrate the latter point, we wanted to visualize the effect of the sub-space projection on the word representation space. Therefore, we used~\newcite{maaten2008visualizing} T-SNE algorithm to project the embeddings into two-dimensions and plotted the words from \textsl{Sem-Lex}, colored according to their sentiment score. We first used the unsupervised embeddings and then, leveraging a sub-space projection (trained on \textsl{Sem-Lex}), induced and plotted new embeddings for the same words. These two plots are shown in Figure \ref{fig:sentiment_emb}. On the left, we can see that unsupervised embeddings can naturally capture sentiment information---words with similar sentiment scores tend to be closer to each other. On the right, we see that in the space induced by the sub-space projection, not only are similar words (w.r.t to sentiment) drawn even closer but also, quite interestingly, the words become arranged in what seems to be a \textit{continuum} from the most negative to the most positive sentiment polarity.

\begin{figure}[htb]
\centering
\includegraphics[width=1\linewidth]{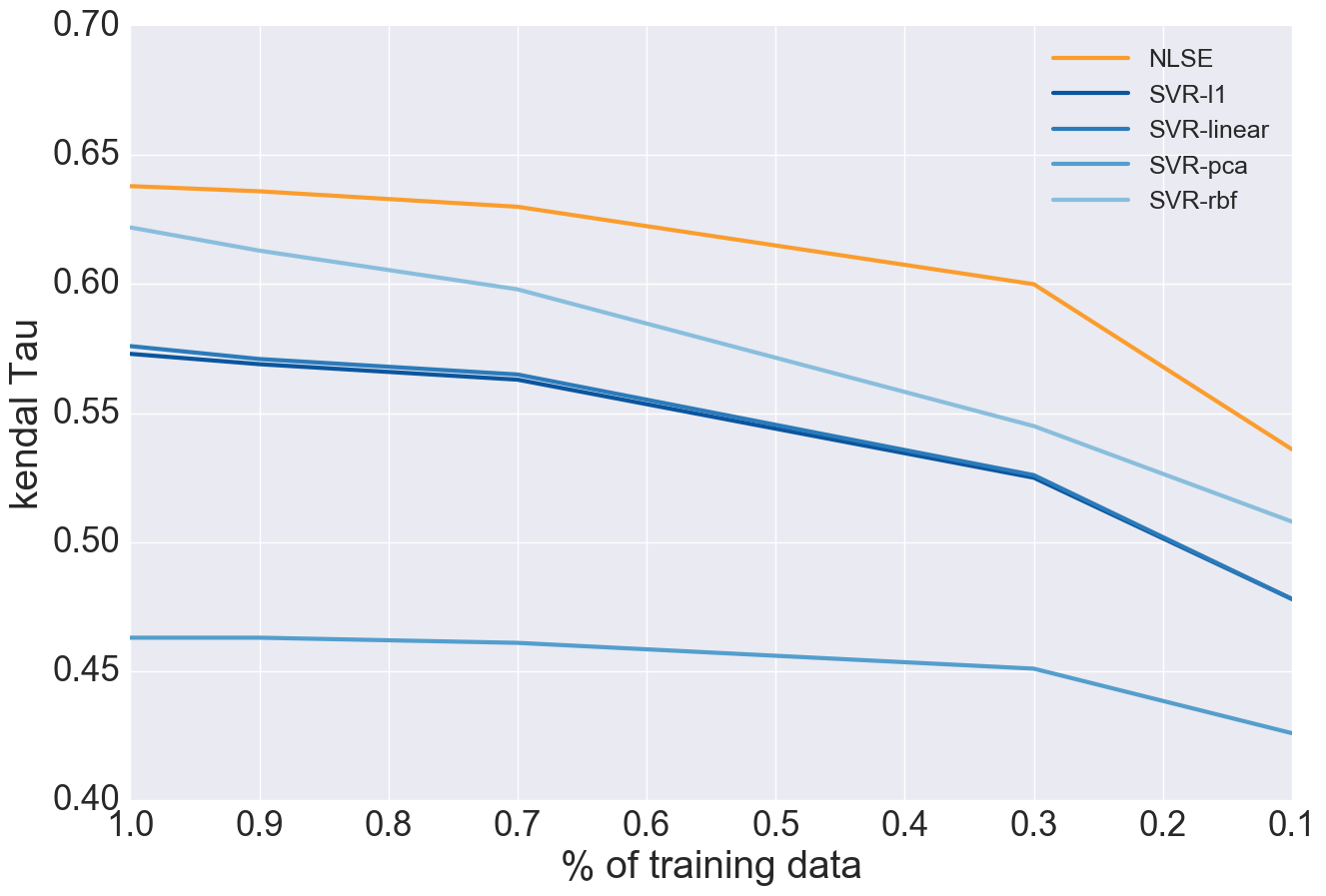}
\caption{Performance of the different baselines in predicting the \textit{happiness score} of words, as a function of the size of the training data.}
\label{fig:fewdata}
\end{figure}

The overall results show that pre-trained word embeddings can indeed capture a wide range of semantic properties, and be leveraged to induce subjective lexicons. Furthermore, the simplicity of our method suggests that it could be used to derive specific lexicons for different domains or demographics, to reflect the fact that some words are used with different connotations by different groups of people~\cite{yang2015putting}. However, this would require creating multiple `training' lexicons, which raises the question of how much data is required to induce high-quality lexicons. To investigate this question, we plotted the performance of the different models, as a function of the training data size (Figure \ref{fig:fewdata}). As expected, the performance of all the models monotonically decreases with less training data. Nevertheless, we observe that the performance of our model decays slower than the RBF baseline (the second best method). Notably, when trained with 30\% of the data, our model attains the same performance of the RBF baseline trained with 70\% of the data.

\section{Lexicon Based Twitter Sentiment Analysis}

\begin{figure*}[tb]
\centering
\begin{subfigure}{.49\linewidth}
  \centering
  \includegraphics[width=1\linewidth]{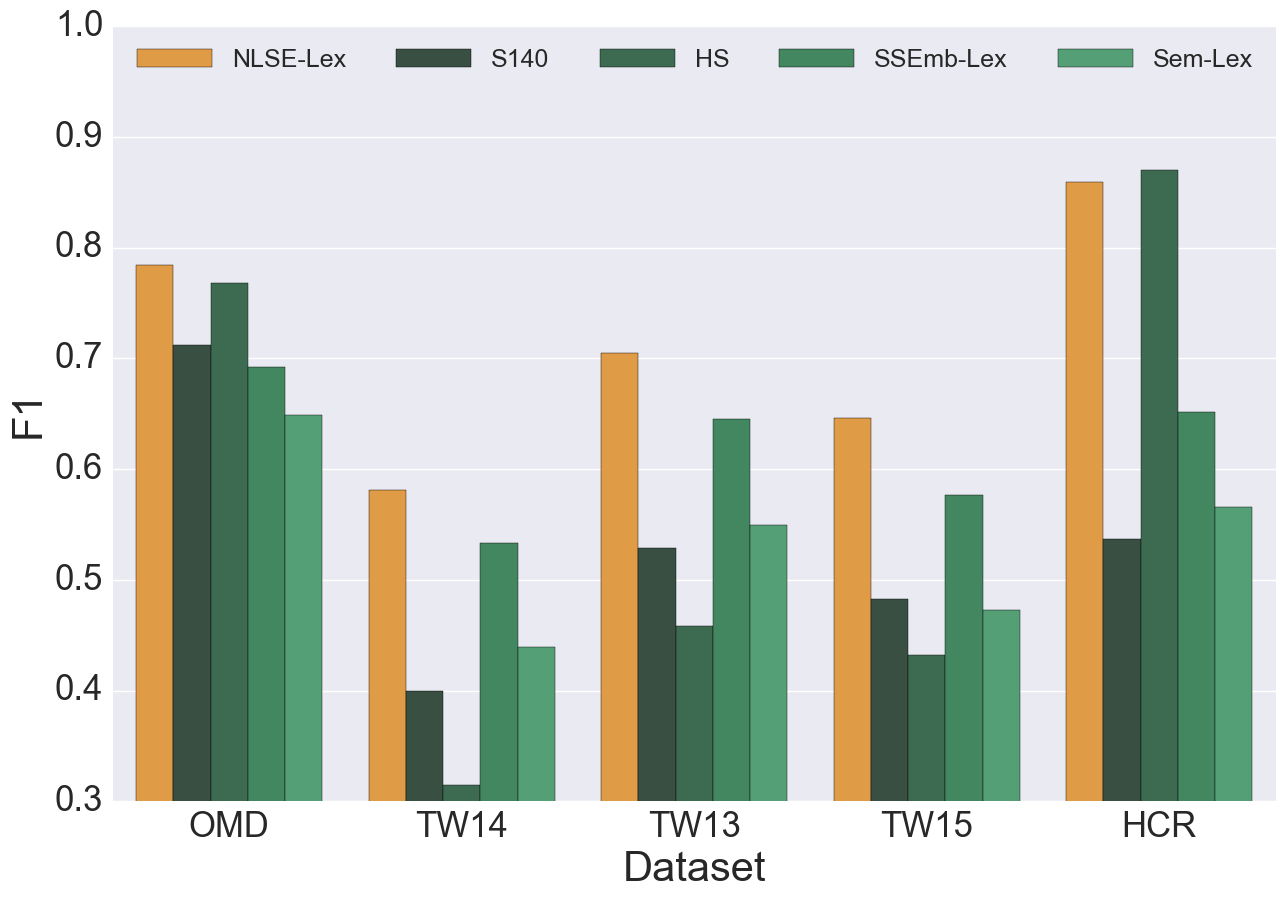}
  \subcaption{Performance of lexicon-based classifiers built on top of different lexicons}
\label{fig:lexicons}
\end{subfigure}%
\hfill
\begin{subfigure}{.49\linewidth}
  \centering
\includegraphics[width=1\linewidth]{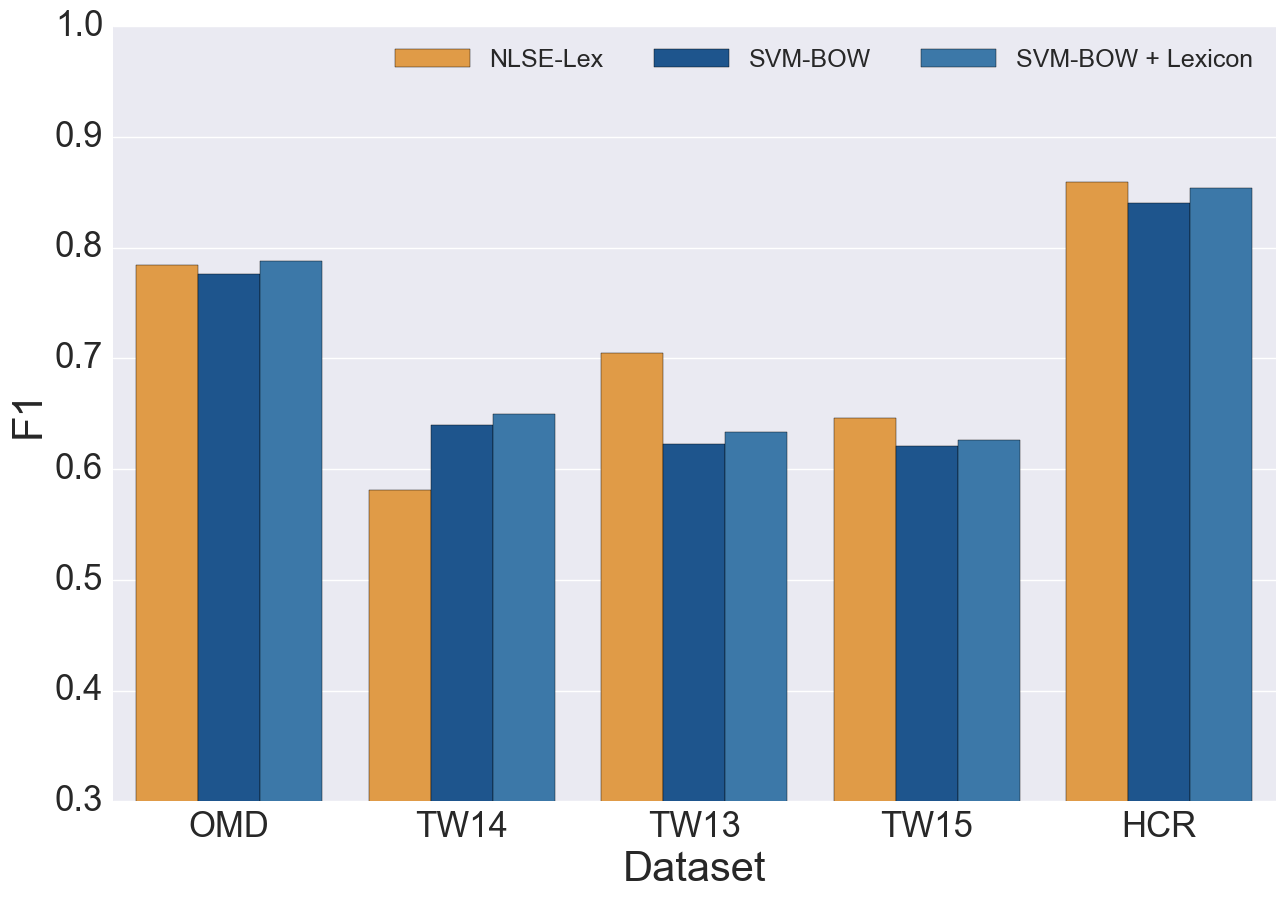}
  \subcaption{Comparison of lexicon-based classifiers against supervised models.}
  \label{fig:lex_vs_svm}
\end{subfigure}
\caption{Results of the sentiment classification experiments}
\label{fig:classification}
\end{figure*}

We now report on a set of experiments designed to assess the quality of our lexicons in downstream applications. To this end, we induced a large-scale sentiment lexicon, henceforth denoted as \textsl{NLSE-Lex}, with a model trained on the \textsl{Sem-Lex} lexicon. Then, we developed lexicon-based sentiment classifiers that infer the polarity of messages by aggregating the \textit{sentiment scores} of individual words. More formally, given a message ${m=\{w_1,\ldots,w_n\}}$ with $n$ words, the overall sentiment is:
\begin{align}
    \text{sentiment}(m;t) &= 
\begin{cases}
    positive, & \text{if } \text{score}(m) \geq t\\
    negative,              & \text{otherwise}
\end{cases} \\
		\text{score}(m) &= \frac{1}{n}\sum_{w_i \in m}y_i
\end{align}
\noindent where, $y_i$ is the sentiment score associated to word $w_i$ in a given lexicon and $t$ is the threshold that separates the positive and negative classes.

We compared the performance of lexicon-based classifiers built on top of the following Twitter lexicons: 

\begin{itemize}
\item \textsl{Sem-Lex}, a small manually labeled lexicon. This will provide a baseline performance;
\item \textsl{NLSE-Lex}, induced with our method;
\item \textsl{Sentiment140 (S140)} and \textsl{Hashtag Sentiment (HL)}, created using term co-occurrence statistics collected from large corpora~\cite{kiritchenko2014sentiment};
\item \textsl{SSEmb-Lex}, obtained using sentiment-specific embeddings~\cite{tang-EtAl2014}.
\end{itemize}

For simplicity, we converted the labels of all the lexicons to the range $[-1, 1]$ and  discarded words with scores between $[-0.2, 0.2]$ to keep only terms that strongly convey sentiment, as suggested by \newcite{dodds2011temporal}. The classification threshold was set with a simple heuristic: we assume that most Twitter posts do not convey any particular sentiment, thus we set the threshold to ${t=\mathbb{E}[\text{score}(m)]}$. In other words, if the score of a message is above the \textit{expected} sentiment score, then it is considered \textit{positive}, otherwise it is \textit{negative}. Finally, we compared the performance of the lexicon-based classifier to that of supervised approaches. For each test set, we trained two SVM models. One using only bag-of-words (\textsl{SVM-BOW}) features; and another, combining BOW features with a set of features extracted from the manually created lexicon: the \textit{mean}, \textit{sum}, \textit{maximum}, \textit{minimum} and \textit{standard deviation} of the word sentiment scores (\textsl{SVM-BOW + Lexicon}).

\begin{table}[htb]
\begin{center}

\begin{tabular}{l r r}
& {\bf \# Training Tweets} & {\bf \# Test Tweets}\\ 
\hline\hline
\textsl{TW-train}  & 6,013 & - \\ 
\textsl{TW13}  & - &  2,173 \\ 
\textsl{TW14}  & - &  1,183 \\ 
\textsl{TW15}  & - &  1,402 \\ 
\hline
OMD  & 1,306 & 598 \\ 
HCR  & 1,257 & 665 \\ 
\hline
\end{tabular}

\end{center}
\caption{Summary of the datasets used in the sentiment classification experiments. The top rows correspond to the test sets from SemEval's Twitter Sentiment Analysis competition; the \textsl{TW-train} dataset was only used as training data. The bottom rows, correspond to the datasets introduced by \newcite{speriosu2011twitter}.}
\label{table:testsets}
\end{table}

The classifiers were evaluated on the following five datasets, summarized in Table \ref{table:testsets}. Three datasets compiled by SemEval for their well-known Twitter sentiment analysis challenge~\cite{SemEval15Task10} (\textsl{TW-13}, \textsl{TW-14} and \textsl{TW-15}) (top rows); and two datasets introduced by \newcite{speriosu2011twitter}---\textsl{OMD}, with reactions to the 2008 USA presidential debate opposing the democrat candidate Barack Obama and republican candidate Jonh Mccain; and \textsl{HCR}, with tweets discussing the 2010 health care reform in the USA. It should be noted that, \citeauthor{speriosu2011twitter} datasets have standard splits for training, development and testing, hence for ease of comparison, our classifiers were evaluated on the test sets. Furthermore, all the aforementioned datasets are labeled in terms of three classes---positive, negative and neutral, but in these experiments we excluded the neutral class and focused on binary classification.

\subsection{Results}

The results of the sentiment classification experiments are presented in Figure \ref{fig:classification}. In Figure \ref{fig:lexicons}, we compare the performance of the different lexicons over the test data. We observe that our lexicon outperform the others in nearly all cases, with the exception of the \textsl{HCR} dataset where the \textsl{HL} lexicon performs marginally better. However, note that this same lexicon obtains the worst performance on \textsl{TW14}. In Figure \ref{fig:lex_vs_svm}, we compare the \textsl{NLSE-Lex} with the supervised models. We found that our lexicon-based classifier is extremely competitive and, somewhat surprisingly, even outperforms the supervised baselines in almost all of the datasets.

\section{Conclusions}

This paper presented a novel approach to induce large-scale subjective lexicons suitable for social media analysis. We exploit the fact that unsupervised word embeddings capture semantic properties of words, and can be used as features for lexicon expansion models. However, instead of using the embeddings directly, we induce and exploit task-specific representations, via sub-space projection. To this end, we leverage the \newcite{astudillo-EtAl:2015:ACL-IJCNLP} NLSE model to jointly learn the adapted representations and respective predictor. The experimental results show that our approach outperforms previous work and other related baselines, across multiple lexicons and subjective properties. Working with lower-dimensional representations also allows us to induce predictors with less training data. Indeed, the results demonstrate that, compared to the other baselines, our method can make better use of limited amounts of training data. Finally, we empirically showed how the sub-space projections learned by the NLSE, transform the embedding space to better capture task-specific information.

To assess the quality of our lexicons, first, we compared the performance of lexicon-based sentiment classifiers built on top of ours, and other large-scale Twitter lexicons. We observed that the classifiers built with our lexicons largely outperform the other baselines. Second, we compared our lexicon-based classifier with supervised models and, surprisingly, we found that our lexicon-based model outperforms the more sophisticated models. These results demonstrate the quality of our lexicon and suggest that, with the appropriate lexicons, simple studies (e.g., involving binary sentiment classification) can be performed without the hassle of creating labeled data.

\bibliographystyle{aaai}
\bibliography{lex}

\begin{thebibliography}{}

\bibitem[\protect\citeauthoryear{Amir \bgroup et al\mbox.\egroup }{2015}]{Amir}
Amir, S.; Ling, W.; Astudillo, R.; Martins, B.; Silva, M.~J.; and Trancoso, I.
\newblock 2015.
\newblock {INESC-ID}: A regression model for large scale twitter sentiment
  lexicon induction.
\newblock In {\em Proceedings of the 9th International Workshop on Semantic
  Evaluation (SemEval 2015)},  613--618.

\bibitem[\protect\citeauthoryear{Astudillo \bgroup et al\mbox.\egroup
  }{2015}]{astudillo-EtAl:2015:ACL-IJCNLP}
Astudillo, R.; Amir, S.; Ling, W.; Silva, M.; and Trancoso, I.
\newblock 2015.
\newblock Learning word representations from scarce and noisy data with
  embedding subspaces.
\newblock In {\em Proceedings of the 53rd Annual Meeting of the Association for
  Computational Linguistics and the 7th International Joint Conference on
  Natural Language Processing (Volume 1: Long Papers)},  1074--1084.

\bibitem[\protect\citeauthoryear{Baroni, Dinu, and
  Kruszewski}{2014}]{Baroni14Dont}
Baroni, M.; Dinu, G.; and Kruszewski, G.
\newblock 2014.
\newblock Don’t count, predict! a systematic comparison of context-counting
  vs. context-predicting semantic vectors.
\newblock In {\em Proceedings of the Annual Meeting of the Association for
  Computational Linguistics}.

\bibitem[\protect\citeauthoryear{Bestgen and Vincze}{2012}]{Bestgen12Checking}
Bestgen, Y., and Vincze, N.
\newblock 2012.
\newblock Checking and bootstrapping lexical norms by means of word similarity
  indexes.
\newblock {\em Behavior Research Methods} 44(4).

\bibitem[\protect\citeauthoryear{Bradley and Lang}{1999}]{bradley1999affective}
Bradley, M.~M., and Lang, P.~J.
\newblock 1999.
\newblock Affective norms for english words (anew): Instruction manual and
  affective ratings.
\newblock Technical report, Citeseer.

\bibitem[\protect\citeauthoryear{Dodds \bgroup et al\mbox.\egroup
  }{2011}]{dodds2011temporal}
Dodds, P.~S.; Harris, K.~D.; Kloumann, I.~M.; Bliss, C.~A.; and Danforth, C.~M.
\newblock 2011.
\newblock Temporal patterns of happiness and information in a global social
  network: Hedonometrics and twitter.
\newblock {\em PloS one} 6(12):e26752.

\bibitem[\protect\citeauthoryear{Esuli and
  Sebastiani}{2006}]{esuli2006sentiwordnet}
Esuli, A., and Sebastiani, F.
\newblock 2006.
\newblock Sentiwordnet: A publicly available lexical resource for opinion
  mining.
\newblock In {\em Proceedings of LREC}, volume~6,  417--422.
\newblock Citeseer.

\bibitem[\protect\citeauthoryear{Harris}{1954}]{harris1954distributional}
Harris, Z.~S.
\newblock 1954.
\newblock Distributional structure.
\newblock {\em Word} 10(2-3):146--162.

\bibitem[\protect\citeauthoryear{Hinton and
  Salakhutdinov}{2006}]{hinton2006reducing}
Hinton, G.~E., and Salakhutdinov, R.~R.
\newblock 2006.
\newblock Reducing the dimensionality of data with neural networks.
\newblock {\em Science} 313(5786):504--507.

\bibitem[\protect\citeauthoryear{Hinton}{1986}]{hinton1986learning}
Hinton, G.~E.
\newblock 1986.
\newblock Learning distributed representations of concepts.
\newblock In {\em Proceedings of the eighth annual conference of the cognitive
  science society}, volume~1, ~12.
\newblock Amherst, MA.

\bibitem[\protect\citeauthoryear{Hu and Liu}{2004}]{hu2004mining}
Hu, M., and Liu, B.
\newblock 2004.
\newblock Mining and summarizing customer reviews.
\newblock In {\em Proceedings of the tenth ACM SIGKDD International Conference
  on Knowledge discovery and data mining},  168--177.

\bibitem[\protect\citeauthoryear{Kamps \bgroup et al\mbox.\egroup
  }{2004}]{kamps2004using}
Kamps, J.; Marx, M.; Mokken, R.~J.; and de~Rijke, M.
\newblock 2004.
\newblock Using wordnet to measure semantic orientations of adjectives.
\newblock In {\em Proceedings of 4th International Conference on Language
  Resources and Evaluation, Vol IV,},  1115--1118.

\bibitem[\protect\citeauthoryear{Kim and Hovy}{2006}]{kim2006identifying}
Kim, S.-M., and Hovy, E.
\newblock 2006.
\newblock Identifying and analyzing judgment opinions.
\newblock In {\em Proceedings of the Human Language Technology Conference of
  the NAACL, Main Conference},  200--207.

\bibitem[\protect\citeauthoryear{Kiritchenko, Zhu, and
  Mohammad}{2014}]{kiritchenko2014sentiment}
Kiritchenko, S.; Zhu, X.; and Mohammad, S.~M.
\newblock 2014.
\newblock Sentiment analysis of short informal texts.
\newblock {\em Journal of Artificial Intelligence Research}  723--762.

\bibitem[\protect\citeauthoryear{Labutov and Lipson}{2013}]{labutov2013re}
Labutov, I., and Lipson, H.
\newblock 2013.
\newblock Re-embedding words.
\newblock In {\em Proceedings of the 51st annual meeting of the ACL},
  489--493.

\bibitem[\protect\citeauthoryear{Ling \bgroup et al\mbox.\egroup
  }{2015}]{ling-EtAl:2015:EMNLP2}
Ling, W.; Dyer, C.; Black, A.~W.; Trancoso, I.; Fermandez, R.; Amir, S.;
  Marujo, L.; and Luis, T.
\newblock 2015.
\newblock Finding function in form: Compositional character models for open
  vocabulary word representation.
\newblock In {\em Proceedings of the 2015 Conference on Empirical Methods in
  Natural Language Processing},  1520--1530.

\bibitem[\protect\citeauthoryear{Maaten and
  Hinton}{2008}]{maaten2008visualizing}
Maaten, L. v.~d., and Hinton, G.
\newblock 2008.
\newblock Visualizing data using t-sne.
\newblock {\em Journal of Machine Learning Research} 9(Nov):2579--2605.

\bibitem[\protect\citeauthoryear{Mitchell \bgroup et al\mbox.\egroup
  }{2013}]{Mitchell2013}
Mitchell, L.; Harris, K.~D.; Frank, M.~R.; Dodds, P.~S.; and Danforth, C.~M.
\newblock 2013.
\newblock The geography of happiness: connecting twitter sentiment and
  expression, demographics, and objective characteristics of place.
\newblock {\em PLoS ONE} 8(5).

\bibitem[\protect\citeauthoryear{Mohammad and Turney}{2013}]{mohammad2013a}
Mohammad, S.~M., and Turney, P.~D.
\newblock 2013.
\newblock Crowdsourcing a word--emotion association lexicon.
\newblock {\em Computational Intelligence} 29(3).

\bibitem[\protect\citeauthoryear{O'Connor \bgroup et al\mbox.\egroup
  }{2010}]{O'Connor2010}
O'Connor, B.; Balasubramanyan, R.; Routledge, B.~R.; and Smith, N.~A.
\newblock 2010.
\newblock From tweets to polls: Linking text sentiment to public opinion time
  series.
\newblock In {\em Proceedings of the 4th International AAAI Conference on
  Weblogs and Social Media}.

\bibitem[\protect\citeauthoryear{Pennington, Socher, and
  Manning}{2014}]{pennington2014glove}
Pennington, J.; Socher, R.; and Manning, C.~D.
\newblock 2014.
\newblock Glove: Global vectors for word representation.
\newblock {\em Proceedings of the 2014 Empiricial Methods in Natural Language
  Processing} 12.

\bibitem[\protect\citeauthoryear{Plutchik}{1980}]{Plutchik}
Plutchik, R.
\newblock 1980.
\newblock {\em {A general psychoevolutionary theory of emotion}}.
\newblock Academic press.
\newblock  3--33.

\bibitem[\protect\citeauthoryear{Rao and Ravichandran}{2009}]{rao2009semi}
Rao, D., and Ravichandran, D.
\newblock 2009.
\newblock Semi-supervised polarity lexicon induction.
\newblock In {\em Proceedings of the 12th Conference of the European Chapter of
  the Association for Computational Linguistics},  675--682.

\bibitem[\protect\citeauthoryear{Rosenthal \bgroup et al\mbox.\egroup
  }{2015}]{SemEval15Task10}
Rosenthal, S.; Nakov, P.; Kiritchenko, S.; Mohammad, S.~M.; Ritter, A.; and
  Stoyanov, V.
\newblock 2015.
\newblock Semeval-2015 task 10: Sentiment analysis in twitter.
\newblock In {\em Proceedings of the 9th International Workshop on Semantic
  Evaluation}, SemEval '2015.
\newblock Denver, Colorado: Association for Computational Linguistics.

\bibitem[\protect\citeauthoryear{Rothe, Ebert, and
  Sch{\"u}tze}{2016}]{rothe2016ultradense}
Rothe, S.; Ebert, S.; and Sch{\"u}tze, H.
\newblock 2016.
\newblock Ultradense word embeddings by orthogonal transformation.
\newblock {\em arXiv preprint arXiv:1602.07572}.

\bibitem[\protect\citeauthoryear{Speriosu \bgroup et al\mbox.\egroup
  }{2011}]{speriosu2011twitter}
Speriosu, M.; Sudan, N.; Upadhyay, S.; and Baldridge, J.
\newblock 2011.
\newblock Twitter polarity classification with label propagation over lexical
  links and the follower graph.
\newblock In {\em Proceedings of the First workshop on Unsupervised Learning in
  NLP},  53--63.
\newblock Association for Computational Linguistics.

\bibitem[\protect\citeauthoryear{Tang \bgroup et al\mbox.\egroup
  }{2014}]{tang-EtAl2014}
Tang, D.; Wei, F.; Qin, B.; Zhou, M.; and Liu, T.
\newblock 2014.
\newblock Building large-scale twitter-specific sentiment lexicon : A
  representation learning approach.
\newblock In {\em Proceedings of the 25th International Conference on
  Computational Linguistics},  172--182.

\bibitem[\protect\citeauthoryear{Tibshirani}{1996}]{tibshirani1996regression}
Tibshirani, R.
\newblock 1996.
\newblock Regression shrinkage and selection via the lasso.
\newblock {\em Journal of the Royal Statistical Society. Series B
  (Methodological)}  267--288.

\bibitem[\protect\citeauthoryear{Tumasjan \bgroup et al\mbox.\egroup
  }{2010}]{tumasjan2010predicting}
Tumasjan, A.; Sprenger, T.~O.; Sandner, P.~G.; and Welpe, I.~M.
\newblock 2010.
\newblock Predicting elections with twitter: What 140 characters reveal about
  political sentiment.
\newblock In {\em 4th International AAAI Conference on Weblogs and Social
  Media}.

\bibitem[\protect\citeauthoryear{Turney and Littman}{2003}]{Turney03Measuring}
Turney, P.~D., and Littman, M.~L.
\newblock 2003.
\newblock Measuring praise and criticism: Inference of semantic orientation
  from association.
\newblock {\em ACM Transactions on Information Systems} 21(4).

\bibitem[\protect\citeauthoryear{Vapnik}{2000}]{vapnik2000nature}
Vapnik, V.
\newblock 2000.
\newblock {\em The nature of statistical learning theory}.
\newblock Springer Science \& Business Media.

\bibitem[\protect\citeauthoryear{Warriner, Kuperman, and
  Brysbaert}{2013}]{warriner2013norms}
Warriner, A.~B.; Kuperman, V.; and Brysbaert, M.
\newblock 2013.
\newblock Norms of valence, arousal, and dominance for 13,915 english lemmas.
\newblock {\em Behavior research methods} 45(4):1191--1207.

\bibitem[\protect\citeauthoryear{Wilson, Wiebe, and
  Hoffmann}{2005}]{wilson2005recognizing}
Wilson, T.; Wiebe, J.; and Hoffmann, P.
\newblock 2005.
\newblock Recognizing contextual polarity in phrase-level sentiment analysis.
\newblock In {\em Proceedings of the conference on human language technology
  and empirical methods in natural language processing},  347--354.
\newblock Association for Computational Linguistics.

\bibitem[\protect\citeauthoryear{Yang and Eisenstein}{2015}]{yang2015putting}
Yang, Y., and Eisenstein, J.
\newblock 2015.
\newblock Putting things in context: Community-specific embedding projections
  for sentiment analysis.
\newblock {\em arXiv preprint arXiv:1511.06052}.

\bibitem[\protect\citeauthoryear{Yu \bgroup et al\mbox.\egroup
  }{2013}]{Yu:2013:UCE:2438098.2438152}
Yu, L.-C.; Wu, J.-L.; Chang, P.-C.; and Chu, H.-S.
\newblock 2013.
\newblock Using a contextual entropy model to expand emotion words and their
  intensity for the sentiment classification of stock market news.
\newblock {\em Knowledge-Based Systems} 41(0).

\end{thebibliography}

\end{document}